\documentclass[nohyperref]{article}

\usepackage{microtype}
\usepackage{graphicx}
\usepackage{subfigure}
\usepackage{booktabs}

\usepackage{hyperref}

\usepackage{amsfonts}
\usepackage{amsmath}
\usepackage{amsthm}
\usepackage[accepted]{icml2022}

\icmltitlerunning{Adversarial Learned Fair Representations using Dampening and Stacking}

\theoremstyle{plain}
\newtheorem{theorem}{Theorem}[section]

\theoremstyle{definition}
\newtheorem{definition}[theorem]{Definition}

\theoremstyle{remark}

\newcommand{\indep}{\perp}

\begin{document}

\twocolumn[
\icmltitle{Adversarial Learned Fair Representations using Dampening and Stacking}

\icmlsetsymbol{equal}{*}

\begin{icmlauthorlist}
\icmlauthor{Max Knobbout}{je}
\end{icmlauthorlist}

\icmlaffiliation{je}{Just Eat Takeaway.com}
\icmlcorrespondingauthor{Max Knobbout}{mknobbout@gmail.com}

\icmlkeywords{Machine Learning, ICML}

\vskip 0.3in
]

\printAffiliationsAndNotice{}

\begin{abstract}
As more decisions in our daily life become automated, the need  to have machine learning algorithms that make fair decisions increases. In fair representation learning we
are tasked with finding a suitable representation of the data in which a sensitive variable is censored. Recent work aims to learn fair representations through adversarial learning. This paper builds upon this work by introducing a novel algorithm which uses dampening and stacking to learn adversarial fair representations. Results show that that our algorithm improves upon earlier work in both censoring and reconstruction.
\end{abstract}
\section{Introduction}
The need to have machine learning algorithms that make fair decisions becomes increasingly important in modern society. A decision is fair if it does not depend on a sensitive variable such as gender, race, or age. Models trained with biased data can lead to unfair decisions \cite{10.1145/3457607}. In fair representation learning we are tasked with finding a suitable representation of the data in which the sensitive variable is censored. This ensures that these representations can be used for any downstream task, such as classification or segmentation, which should not rely on the value of the sensitive variable \cite{pmlr-v28-zemel13}. Throughout this paper, we often refer to this sensitive variable as the \emph{protected variable}.

It is important to note that the notion of fairness is not trivial, and a multitude of fairness constraints have been proposed \cite{10.1145/2090236.2090255}. This paper in particular attempts to learn representations such that any predictor trained on these representations abide to \emph{statistical parity} with regards to the protected variable. This means that the classifier should treat the data containing the protected variable statistically similar to the general population, i.e. there is no bias towards the protected variable.

An additional point on fair representation learning is that it is not always the case that (un)fairness is our main concern. Often times, our representations contain some \emph{intended} but \emph{unwanted} bias which we want to eliminate. Suppose we are given a set of documents, the task is to discover the topics that are present. Moreover, once we find the topics, we want to associate with each of the topics a sentiment score (either positive or negative) by computing the sentiment of each individual document. Traditional algorithms use the underlying word distribution as a basis for a topic model. With the rise of more advanced language models such as BERT \cite{devlin-etal-2019-bert}, which are very proficient at modelling semantic similarity of sentences across multiple languages, it is no surprise that these models are also used for topic mining \cite{DBLP:journals/corr/abs-2103-00498}. These models are often pretrained on huge text corpora and the resulting sentence embeddings contain a lot of information, even about the sentiment of the sentence. In other words, if we would use these representations together with a clustering algorithm to find the topics, what we will often find is that some topics are centered around sentiment, i.e. a topic only containing ``positive'' sentences. Since it is in most cases infeasible to retrain these models in order to remove the unwanted bias, we are forced to remove the unwanted bias afterwards from the representations. With the increasing popularity of pretrained models for transfer learning across a wide variety of tasks related to audio, speech, language, and images, the need for a strong algorithm for learning fair unbiased representations becomes apparent.

Often with learning a fair representation, the naive approach of dropping certain features of the data is insufficient. The origin of the bias might latently depend on some nonlinear combination of other variables, and can thus leak back into a decision making model. This inspired the work by \cite{edwards2016censoring}, which aims to learn a fair representation through adversarial learning. They use an auto-encoder as a generator for the new representation whose aim is to learn a new latent representation which attempts to censor the protected variable for the adversary. This work was later extended in \cite{pmlr-v80-madras18a} where they propose learning objectives for other fairness metrics such as equalized odds and equal opportunity. In \cite{kenfack2021adversarial} this work was further extended by introducing stacked auto-encoders to enforce fairness and improve censoring at different latent spaces.

This work builds on the previous adversarial approach. In particular it focuses on the case where the downstream task we may encounter is unknown, i.e. it can be either some supervised classification objective or some unsupervised clustering or segmentation objective. The challenge with learning fair representations is that on one hand we want to censor the data, and on the other we want to retain as much information as possible. Since these objectives are often opposed, the approaches in \cite{edwards2016censoring}, \cite{pmlr-v80-madras18a}, \cite{kenfack2021adversarial}, and various others define the global objective of the model as a weighted sum of reconstruction error and predictive loss. This requires the trainer of a model to select some suitable hyperparameter which defines how much we value reconstruction error over predictive loss. This hyperparameter often has a large impact on the learned representations we get, and we can identify at least three issues with it. Firstly, we have no a priori knowledge on how the reconstruction error and the predictive loss relate. It could be nonlinear, which makes it almost impossible to make an informed decision beforehand. Secondly, the value of this hyperparameter gives us no formal guarantee of the censoring capabilities of the model. Some values can cause a collapse of the model. Thirdly, the hyperparameter choice is not explainable to the relevant stakeholders of the model. This makes it impractical for most industry use cases where hyperparameter choices need to be justified. As such, many authors using this methodology such as \cite{edwards2016censoring}, \cite{pmlr-v80-madras18a}, \cite{Feng2019LearningFR}, and \cite{kenfack2021adversarial} either use a trial-and-error approach, or an arbitrary chosen constant, with regard to the choice of this hyperparameter. More often than not, the censoring capabilities of the learned representation are a hard constraint of the model. Thus, in many industry use cases, we are only interested in finding solutions in some restricted hypothesis space abiding some censoring constraint.

A second perhaps even greater issue with the previous work is its instability. In particular, due to the unstable dynamic between actor and adversary we often learn suboptimal solutions. This has been observed in many cases such as \cite{edwards2016censoring} and \cite{kenfack2021adversarial}, but never fully addressed.

This paper attempts to mitigate these issues by introducing a novel algorithm for learning fair representations. In particular, it uses dampening to stabilize the interaction between actor and adversary, and uses stacking to learn strong censored representations within a restricted hypothesis space. As we will see, this algorithm outperforms the current approach in both censoring and reconstruction.

The remainder of the paper is structured in the following way: in Section~\ref{related_work} we briefly reiterate related work, in Section~\ref{problem_definition} we formally define the problem, in Section~\ref{our_approach} we introduce our approach and novel algorithm, in Section~\ref{experiments} and \ref{results} we discuss the experiments and results, and in Section~\ref{conclusions} we conclude this work.
\section{Related Work}
\label{related_work}
In \cite{pmlr-v28-zemel13} the first fair representation learning approach was presented. Their methodology aims to map input data to a new representation in terms of a probabilistic mapping to a set of prototypes. 

In \cite{DBLP:journals/corr/LouizosSLWZ15} an architecture based on the Variational Auto-Encoder was proposed in order to learn fair representations, called the Variational Fair Auto-Encoder. This model uses priors to encourage independence between the sensitive and latent factors of variation.

In \cite{edwards2016censoring} the first adversarial approach was introduced to learning fair representations. They use an auto-encoder as a generator for the new representation whose aim is to learn a new latent representation which attempts to censor the protected variable for the adversary. This work was later extended by  \cite{pmlr-v80-madras18a} where they propose learning objectives for other fairness metrics. In \cite{kenfack2021adversarial} this work was further extended by introducing stacked auto-encoders to enforce fairness at different latent spaces.

This work builds and improves upon this previous work on adversarial learning. We propose a novel algorithm using dampening and stacking for learning fair representations which increases censoring and decreases reconstruction error. 

\section{Problem Definition}
\label{problem_definition}
This paper focuses purely on representation learning rather than classification. Our aim is to learn a fair representation independent of the downstream task we may encounter (supervised or unsupervised). We adopt the notation of \cite{edwards2016censoring} of using the letter $X$ to represent the data, and $S$ to represent the protected variable. Each $x_i \in X$ is assumed to be some real-valued vector $x_i \in \mathbb{R}^n$, and each $s_i \in S$ is either 0 or 1, denoting if instance $i$ is sensitive or not: $s_i \in \{0,1\}$. We adopt the following fairness criteria: given data $X$ and protected variable $S$, we aim to learn a new representation $f(X)$ for which it holds that for any predictor $g$ derived from $f(X)$ we have $g(f(X)) \indep S$, i.e. $g(f(x))$ and $S$ are independent, often denoted as the fairness constraint of \emph{statistical parity} \cite{10.1145/2090236.2090255}. In short, our aim is to find a representation $f(X)$ which give no predictive preference towards $S$. Throughout this paper we refer to $f(X)$ as the \emph{censored} representation. It is important to note that the censored representation is not (necessarily) in the same space as the original data, and can have a different number of dimensions.

On one hand we are aiming to censor the representation, while on the other hand our goal is to retain as much information as possible. In order to capture these opposite objectives, we frame our setting as an adversarial learning problem. Similar to \cite{edwards2016censoring} and various papers following this, we model two agents with competing objectives:
\begin{itemize}
\item An auto-encoder $e$ with corresponding decoder $d$ representing the actor; and 
\item A classifier $h$ representing the adversary.
\end{itemize}
As per usual, $e$, $d$ and $h$ are implemented in this paper using a feed-forward neural network. We aim to find a censored representation $e(X)$. The objective of the adversary is to predict $S$ from the censored representation $e(X)$, while the aim of the actor is to learn this censored representation such that $d(e(X))$ is as close to $X$ as possible (the normal objective of an auto-encoder) \emph{and} to deny the adversary from being able to learn $S$ from $e(X)$.

To make these notions precise, let $\mathcal{L}^{\mathop{act}}_{e,d}$ be the loss of the auto-encoder. We set this loss to be the mean-squared error (MSE), otherwise known as the reconstruction error:
\[\mathcal{L}^{\mathop{act}}_{e,d} = \frac{1}{|X|}\sum_{x_i \in X}\left\lVert x_i - d(e(x_i))\right\rVert_2^2\]
Moreover, we set the loss of the adversary to be the \emph{negative} cross-entropy loss over $S$:
\[
\mathcal{L}^{\mathop{adv}}_{e,h} = \frac{1}{|X|} \sum_{s_i, \hat{s_i} \in S, h(e(X))} s_i \mathop{log}(\hat{s_i}) + (1-s_i) \mathop{log}(1-\hat{s_i})
\]
Note that the above formulation, following the convention of \cite{edwards2016censoring} and \cite{pmlr-v80-madras18a}, takes the \emph{negative} of the usual cross-entropy. Defining the loss as a maximization problem rather than a minimization one, allows us to state a joint objective of the actor and adversary in the form of a min-max problem. Particularly, we let $\mathcal{L}(e,d,h)$ be the joint loss, and we define it  as a weighted sum of $\mathcal{L}^{\mathop{act}}_{e,d}$ and $\mathcal{L}^{\mathop{adv}}_{e,h}$:
\[\mathcal{L}(e,d,h) = \mathcal{L}^{\mathop{act}}_{e,d} + \alpha \mathcal{L}^{\mathop{adv}}_{e,h} \]
Here $\alpha$ is some predetermined chosen hyperparameter denoting the importance of $\mathcal{L}^{\mathop{act}}_{e,d}$ over $\mathcal{L}^{\mathop{adv}}_{e,h}$. Since we consider the negative cross-entropy loss, the aim is to \emph{minimize} this loss under the assumption that the adversary is trying to \emph{maximize} this. Thus, our aim is to find $e$ and $d$ which satisfy the following:
\[\min_{e,d}\max_h\mathcal{L}(e,d,h) \]
Once we have found such an $e$ and $d$, we are usually only interested in the censored representation $e(X)$, which as noted earlier can be embedded in a different space than $X$. However, sometimes we are interested in the \emph{censored original} representation $d(e(X))$, since these representations share a lot of the inherent properties of $X$ both dimension- and structure-wise. Thus, depending on the use case and task, we may choose to use $e(X)$ or $d(e(X))$ as the censored data.

\subsection{Restricting the Hypothesis Space}
\label{hypothesis_space}
A big problem with the join loss function $\mathcal{L}(e,d,h)$ is the correct choice of $\alpha$. This hyperparameter needs to be selected beforehand, and it has a large impact on the representations we learn. Since (1) we have no a priori knowledge on how $\mathcal{L}^{\mathop{act}}_{e,d}$ and $\mathcal{L}^{\mathop{adv}}_{e,h}$ relate for a given $X$ and $S$, (2) choices of $\alpha$ give us no formal guarantee on the censoring ability of the encoder, and (3) a choice of $\alpha$ is hard to explain to the relevant stakeholders, any choice of $\alpha$ is hard to justify and interpret. Additionally, low and high values for $\alpha$ could cause the trivial function to be learned, i.e. either the encoder learns an uncensored representation, or a constant function is learned.

In order to eliminate these problems, we propose a different objective function. More often than not, the censoring capabilities of the target function are a hard constraint on the resulting model. We recognize that perfect censoring is in most cases not feasible, and as such these hard constraints should define a hypothesis space of possible target functions. An example hard constraint could be that we do not wish that a very competent adversary receives above 60\% accuracy on trying to classify the gender based on a loan application. Such a hard constraint solves the problem of not having a formal guarantee of the target function, and is both more intuitive and explainable to the relevant stakeholders.

To formalize this, we assume a score function $\mathop{score}_{X,S,e}(h)$ and accompanying threshold $T$ which evaluates the performance of adversary $h$ based on data $X$, $S$ and encoder $e$. The constrained hypothesis space $\hat{E}$ for $e$ can be defined as follows:

\[ \hat{E} =\{ e \mid \mathop{score}_{X,S,e}(\mathop{\arg\max}_h( \mathcal{L}^{adv}_{e,h})) \leq T \} \]
We assume that $\mathop{score}$ and $T$ are chosen in accordance with the distribution of $S$ such that $\hat{E}$ is nonempty, e.g. if we are interested in accuracy, $T$ should at least be 50\%. The global objective is now to simply to minimize $\mathcal{L}^{\mathop{act}}_{e,d}$ by only considering encoders from the viable hypothesis space $\hat{E}$:
\[ \min_{e \in \hat{E}, d} \mathcal{L}^{\mathop{act}}_{e,d} \]
In order to find solutions in $\hat{E}$, different optimization methods need to be used. One of the main contributions of this paper is that we supply such an algorithm.

\section{Our Approach}
\label{our_approach}
In this section we will work towards a novel algorithm for learning better censored representations. Before delving into the technical details, it is worthwhile to discuss the shortcomings of the current approach. As mentioned in \cite{edwards2016censoring}, \cite{pmlr-v80-madras18a}, and various other papers, it is very difficult to train these models due to the unstable dynamic between the actor and the adversary. This is true for adversarial learning in general because of the underlying saddle point optimization problem. Different approaches in literature have been proposed to stabilize adversarial networks, ranging from simple solutions such as early stopping or weight clipping \cite{pmlr-v70-arjovsky17a} to more intricate ones such as adding extra stabilizing steps \cite{yadav2018stabilizing}. 

In the context of our setting, it is worthwhile to investigate what the root cause of the instability is. During training, the actor is continuously updating in the direction to make the adversary worse at predicting the protected variable (recall that the objective of the actor is to minimize the \emph{negative} cross-entropy loss of the adversary, while the adversary is trying to \emph{maximize} this). A key insight here is that the loss signal that the adversary is giving to the actor is paradoxical in nature:
\begin{itemize}
    \item If the magnitude of the loss is relatively high, the adversary is incompetent at predicting $S$. Since the adversary is also continuously updating its own loss towards 0, it typically means that we are at a point where the gradient of the loss will be relatively high. This in turn will result in a big update of the actor. However, the adversary was already incompetent at predicting $S$, but we are performing a big update in an uninformative direction when we would rather perform a smaller conservative update.
    \item If the magnitude of the loss is relatively low, the adversary is competent at predicting $S$. When the loss is relatively smooth at local maxima, a low magnitude of the loss will more often than not result in a small update of the actor. However, the adversary was already competent at predicting $S$, but we are performing a small update in an informative direction when we would rather perform a bigger less careful update.
\end{itemize}
The key issue is thus that if the adversary is too competent then the gradients will be weak, and if the adversary is too incompetent the gradients will be uninformative. This interplay between the competence of the adversary and the size of the gradients is also mentioned in \cite{edwards2016censoring}, but not further explored. 

Now consider what this means when we are actually training and updating the actor and adversary. When we adjust the weights of the actor and adversary in turns, as described in \cite{NIPS2014_5ca3e9b1}, we run the risk that the adversary will never be sufficiently competent in the task. This is particularly true when we use a strong adversary with a lot of parameters, which typically need more batches to properly converge. This means that we are constantly making big weight adjustments in an imprecise direction, which again causes the adversary to be incompetent. On the other hand when we train the adversary in the inner loop, apart from it being very inefficient, we would also run the risk of the adversary being too strong, and not being able to make any meaningful updates.

To mitigate these problems and to make the training process more stable, we introduce the notion of \emph{dampening} in the next section. Afterwards we will discuss the notion of \emph{stacking}, allowing us to continuously increase the censoring capabilities of the encoder. Dampening, in combination with stacking forms the basis of our novel algorithm.
\subsection{Dampening}
Dampening is a function that will serve as a stabilizing method of our algorithm in the interaction between actor and adversary. As we will see, dampening returns a number between 0 and 1 denoting how much information the classifier has over a training sample. First, let us define $g$ as a function over subsets $S'$ of our protected variable $S' \subseteq S$:
\[
g(S') = \frac{1}{|S'|}\max\left(\sum_{s_i'\in S'} s_i', \sum_{s_i'\in S'} 1- s_i'\right) 
\]
In words, $g(S')$ represents the best possible accuracy a predictor can receive when using only information about $S'$. Observe that since the protected variable is binary, $g(S') \in [0.5,1]$. The role of $g(S')$ is to serve as a baseline guessing accuracy.

Using this guessing accuracy, we can define dampening $d$. In the below definition, we use $\mathop{acc}(f,X',S')$ as shorthand notation to denote the accuracy score of $f$ on training sample $X',S'$.
\begin{definition}[Dampening]Given a classifier $f$ and training sample $X',S'$, dampening $d$ is defined as:
\[
\mathop{d}(f, X', S') = \frac{\max(0, \mathop{acc}(f, X', S') - g(S'))}{1-g(S')}
\]
Whenever $g(S')=1$, we set $\mathop{d}(f,X',S') = 0$.
\end{definition}
In words, dampening $d(f, X', S') \in [0,1]$ tells us the percentage decrease of number of misclassifications would we use $f$ instead of guessing the most frequent label in the sample. Whenever dampening is 1 for $f$, we know that $f$ achieves perfect accuracy on the training sample $X', S'$, and whenever dampening is 0 we would be no worse off by just informed guessing. Thus, dampening is a measure of information a classifier has over a certain classification task. An important property of dampening is that it is contained within the unit interval, meaning that when we use it as a scaling factor the corresponding result will never be larger than the original value. We experimented with different notions of bounded information, such as the standard Pearson correlation and the $\phi$-coefficient (MMC), but found dampening to work the best for a variety of tasks. We suspect it is due to its linear scaling with the number of correctly classified samples whenever its value is nonzero; a small increase in correctly classified samples translates in a small increase in dampening, and vice versa for a big increase.
\subsection{Stacking}
Stacking is a technique for censoring which was recently introduced in \cite{kenfack2021adversarial}. The idea is that during training we start out with a simple encoder which learns a censored representation. After this initial training phase, we freeze the encoder and append a new trainable one. This process continues until we are completely done with training. Another perspective on this process is that once we learn a censored representation, we recursively start over a completely new training process, except that we use the censored representation as the new input. The key idea behind stacking is that once a censored representation is learned and frozen, it is highly likely that \emph{some} information about the protected variable is lost for good. Thus in theory, repeating the stacking operation can give us representation with arbitrary strong censoring properties.

It is important to note that the authors found that stacking did increase censoring over the original approach, but unsurprisingly also comes at the cost of reconstruction error. In other words, stacking should preferably be combined with a very careful and stable censoring algorithm, which in our case is handled by the addition of dampening. Stacking together with dampening serves the basis for our algorithm, which as we will see outperforms the basic approach in both censoring and reconstruction.
\subsection{The Algorithm}
In Algorithm~\ref{alg:alfrds} we propose our new algorithm called ALFR-DS (``Adversarial Learned Fair Representations using Dampening and Stacking''). This algorithm differs from basic ALFR, as discussed in \cite{edwards2016censoring}, on three key aspects:
\begin{itemize}
    \item We introduce a different loss function for the actor and adversary, instead of the same function $\mathcal{L}(e,d,h)$ as given in Section~\ref{problem_definition} for both.
    \item We use an inner loop for normal backpropagation, and add an extra outer loop which incorporates stacking. An extra termination condition is added which allows us to find solutions in the restricted hypothesis space $\bar{E}$, as defined in Section~\ref{hypothesis_space}.
    \item We train the actor and adversary \emph{concurrently} instead of interleaved. 
\end{itemize}
\begin{algorithm}[tb]
   \caption{ALFR-DS}
   \label{alg:alfrds}
\begin{algorithmic}
   \STATE Initialize $e = \mathop{id}$ \COMMENT{Start with the "empty" encoder.}
   \STATE Initialize $\theta^{\mathop{act}}$ and $\theta^{\mathop{adv}}$ randomly.
   \REPEAT
   \STATE Initialize $e_{\mathop{new}}$ randomly.
   \STATE $e = e_{\mathop{new}} \circ e$ \COMMENT{Add new encoder to the (frozen) stack.}
   \REPEAT
   \STATE $X', S' = $ random mini-batch from $X, S$
   \STATE $L^{\mathop{act}} = \mathcal{L}^{\mathop{act}}_{e,d}(X')$
   \STATE $L^{\mathop{adv}} = \mathcal{L}^{\mathop{adv}}_{e,h}(X', S')$
   \STATE $\delta = d(h \circ e, X',S')$
   \STATE $\theta^{\mathop{act}} = \theta^{\mathop{act}} - \eta \cdot \left( \nabla_{\theta^{\mathop{act}}}  L^{\mathop{act}} +  \delta \cdot \nabla_{\theta^{\mathop{act}}} L^{\mathop{adv}} \right)$
   \STATE $\theta^{\mathop{adv}} = \theta^{\mathop{adv}} + \eta \cdot  (1-\delta) \cdot \nabla_{\theta^{\mathop{adv}}} L^{\mathop{adv}}$

   \UNTIL{Sufficient epochs reached.}
   \STATE Freeze encoder $e$.
   \UNTIL{$\mathop{score}_{X,S,e}(h) \leq T$ or Deadline reached.}
\end{algorithmic}
\end{algorithm}

In the description of the algorithm, we use $\mathcal{L}^{\mathop{act}}_{e,d}(X')$ and $\mathcal{L}^{\mathop{adv}}_{e,h}(X',S')$ to denote the loss functions defined in Section~\ref{problem_definition} applied to $X'$ and $S'$. We use $\theta^{\mathop{act}}$ and $\theta^{\mathop{adv}}$ to refer to the model parameters of the actor ($e$ and $d$) and subsequently the adversary ($h$). The fixed learning rate $\eta$ can be replaced with a parameter-dependent dynamic one: in all of our experiments we have found the Adam optimizer to work the best \cite{journals/corr/KingmaB14}. 

The role of dampening in the algorithm is to act like a ``fuzzy'' turn-taking mechanism: whenever the adversary is weak, $\delta$ will be close to $0$ in our algorithm. This means the actor will hardly use the loss of the adversary in updating the censored representations, i.e. it will act like a normal auto-encoder. Since representation learning in a normal auto-encoder is stable, it gives the adversary time to learn and catch up. Whenever the adversary is strong, $\delta$ will be close to $1$, meaning the adversary will hardly update itself. This allows the auto-encoder to incorporate the loss of the adversary and learn a new censored representation. In other words, the actor can catch up. This is how dampening stabilizes the learning process: it gives either the actor or the adversary time to catch up, resulting in only informative updates of the model. In our experiments we found that no extra stabilizing methods such as gradient clipping were needed.

The width and depth of the encoder that is added to the stack can be varied at any moment. In our implementation, every subsequent encoder after the first uses the same input and output size. In order to be able to censor nonlinear relations in the data at every step, we have given every encoder a single hidden layer. It is important to note that the adversary $h$ can have any neural architecture, depending on the desired censoring strength of the resulting model. It is also important to note that a strong adversary typically means that we need to train the model longer.

The termination condition $\mathop{score}_{X,S,e}(h) \leq T$ tells us when an encoder is contained within the desired hypothesis space given adversary $h$. Although not explicitly mentioned in the algorithm, it is sometimes beneficial to fully train $h$ on the training data $X, S$ without additionally training the encoder after termination of the inner loop. This is to ensure that the adversary is fully converged before we make an assessment about the censoring capabilities of the model. Under the reasonable assumption that $\mathop{score}$ is chosen in such a way that it eventually decreases as $\mathcal{L}^{\mathop{adv}}_{e,h}$ decreases, and that $\mathcal{L}^{\mathop{adv}}_{e,h}$ decreases after a completion of the inner loop, we know that $\mathop{score}_{X,S,e}(h) \leq T$ will eventually hold. In other words, our algorithm will eventually find a solution in the constrained hypothesis space. In order to encourage that $\mathcal{L}^{\mathop{adv}}_{e,h}$ decreases after a completion of the inner loop, we can decrease the hidden size of each encoder that we add to the stack, or we can terminate the inner loop early. However, since it is often undesirable that the model complexity of the encoder grows extremely large, an extra deadline criterion is used to allow us to terminate early. Whenever an early termination occurs, a solution with the desired censoring capabilities was not found within the complexity bounds of the model.

\section{Experiments}
\label{experiments}
We present experiments on two standard widely used datasets. For fair representation learning, which is not directly linked to a classification task, there is no de facto benchmark. Our aim for a certain task is to find a model which can get the lowest reconstruction error (leaves as much as the information intact) and makes sure the adversary achieves the lowest possible negative cross-entropy for predicting the sensitive variable. To make our results more interpretable, we chose to measure the latter by measuring the accuracy of the adversary: in this case, \emph{lower} accuracy is \emph{better}. Our results still hold while using different evaluation metrics, such as average loss or F-score. For our benchmarks, we chose to use the Database of Handwritten Images (MNIST) \cite{deng2012mnist} and the Large Movie Review Dataset (ACL-IMDb) \cite{maas-EtAl:2011:ACL-HLT2011}. 

\subsection{Setup}
In our experiments, we compare ALFR-DS to ALFR in its ability to reconstruct and censor. In order to ensure a fair comparison, every model was trained for 30 epochs. For ALFR-DS, we considered 3 variants: \textbf{ALFR-DS(1)} which runs the inner loop of Algorithm~\ref{alg:alfrds} once for 30 epochs, \textbf{ALFR-DS(2)} which runs the inner loop twice for 15 epochs, and \textbf{ALFR-DS(3)} which runs the inner loop three times for 10 epochs. Both ALFR-DS and ALFR used the same Multi-Layer Perceptron (MLP) adversary. However, we observed that ALFR typically performs better against a weak adversary, so we also considered a variant of ALFR against an adversary using simple Logistic Regression (LR). We refer to these two variants as \textbf{ALFR (MLP)} and \textbf{ALFR (LR)}. For ALFR we tried several values for $\alpha$, but we report only for $\alpha = 1$ since ALFR-DS outperforms ALFR for all nontrivial choices of $\alpha$ in \emph{both} censoring and reconstruction. Both ALFR-DS and ALFR were given the same number of target dimensions to embed to. In order to measure how well a model censors, each model was trained on one slice of the data, after which another slice was used for evaluation. Two new classifiers, one using $\textbf{LR}$ and one using an $\textbf{MLP}$, were freshly trained on these new representations and were asked to predict protected variable $S$. The resulting accuracy scores were used as a benchmark. Additionally, we trained a normal auto-encoder that did not use an adversary and called this \textbf{Uncensored} (which is equivalent of using ALFR with $\alpha = 0$). All experiments were repeated 10 times to account for naturally occurring deviations in the results.

\subsection{MNIST}
\begin{figure}[ht]
\vskip 0.2in
\begin{center}
\centerline{\includegraphics[width=\columnwidth]{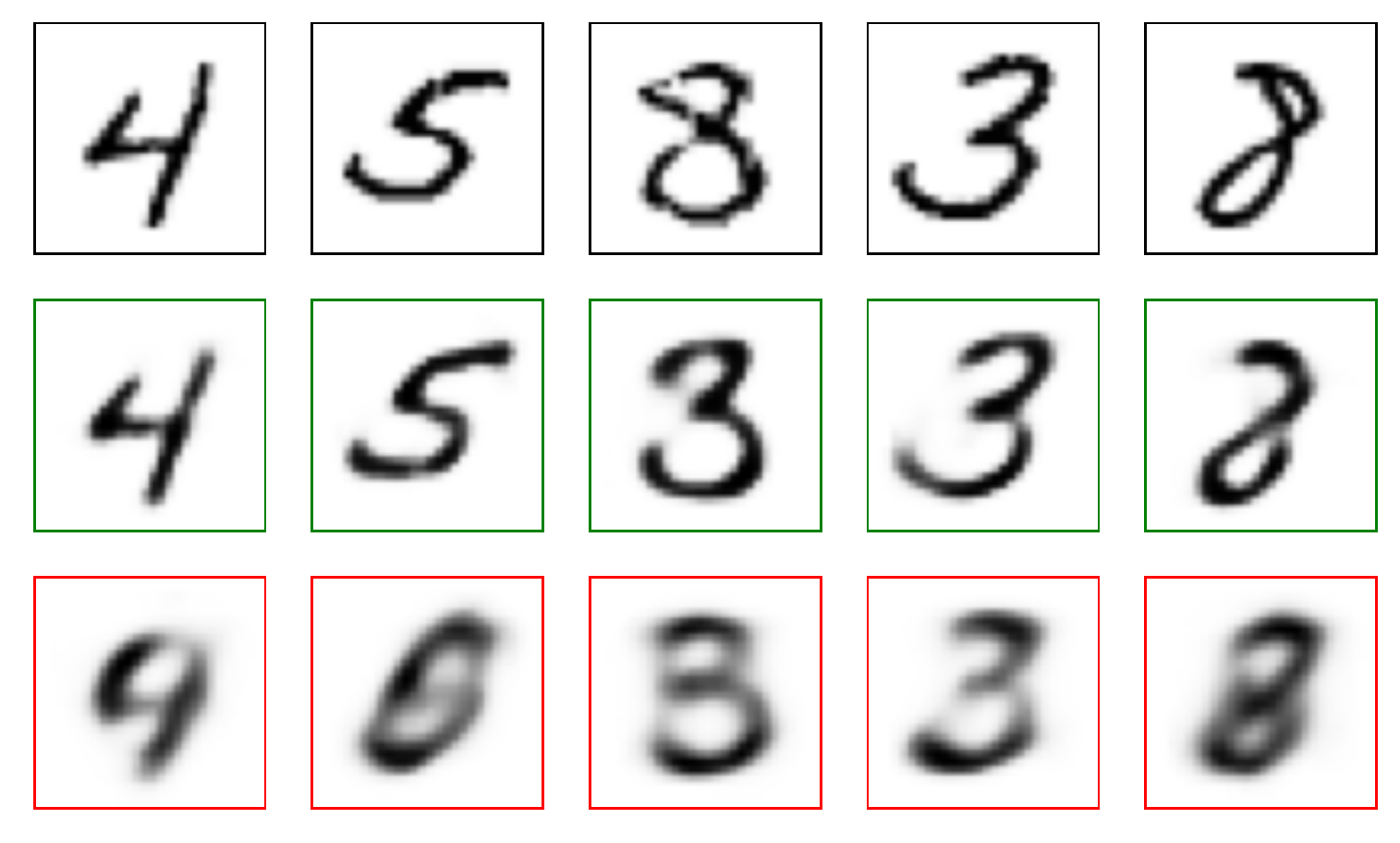}}
\caption{The top row contains the original MNIST digits. The middle row contains the digits censored by our algorithm (ALFR-DS(3)). The bottom row contains the digits censored by the original approach (ALFR (LR)).}
\label{mnist_digits}
\end{center}
\vskip -0.2in
\end{figure}

Motivated by the image anonymization task proposed in \cite{edwards2016censoring}, we propose a censoring task based on the widespread used MNIST dataset of handwritten images. The dataset contains 60,000 examples with corresponding labels, some examples can be seen in Figure~\ref{mnist_digits} in the top row. The goal of this task is to censor all the 8s in the dataset, i.e. we set protected variable $s_i$ for $i$ to 1 whenever the label is 8, and otherwise to 0. Even though this task does not serve a direct practical use, it is a great benchmark for its inherent challenging properties. In particular, the protected variable is \emph{uneven distributed} and the task is very \emph{nonlinear} in nature. It is often not clear how to anonymize a single digit while also keeping the data intact. Moreover, it is a widely spread used dataset, which allows us to easy replicate results. Finally, the censored original space allows us to visually inspect the censoring quality of the network. 

In Figure~\ref{mnist_digits} we get a first sample of the different outputs of ALFR-DS(3) (middle) and ALFR (LR) (bottom). In one instance ALFR-DS censors an 8 to a 3 and in another instance an 8 to a 6. From this visual inspection, it is clear that a lot more information is lost about the original images with ALFR. We tend to see an ``imprint'' of an 8 across most images in the censored representations. Although we suspect that ALFR-DS retains more information, the results are shown in the next section.
\subsection{ACL-IMDb}
Closely related to \cite{DBLP:conf/fat/SweeneyN20} and \cite{kenna2021using} where they perform a debiasing on word embeddings, we are concerned with removing sentiment bias from \emph{sentence} embeddings. Motivated by what we wrote in the introduction, this task is very reminiscent of an actual problem we might encounter, and the main inspiration for our algorithm. The dataset contains 25,000 reviews of movies, which are either labeled positive (1) or negative (0). We set our protected variable $S$ to be equal to this. To get our representations, we embed each review to a 512-dimensional vector using the latest Universal Sentence Encoder for English \cite{cer-etal-2018-universal}. 

In order to get an idea of the importance of censoring sentiment in the context of topic mining, we perform the following data exploration. First we perform K-Means clustering (K=20) on the data, and set these clusters to be the topics. Next, we compute for each topic the top correlated words using Pearson correlation. For each topic, we use the top 3 correlated words as a topic description. The result is shown in  Table~\ref{imdb_topics} for both the original and censored representations using ALFR-DS.
\begin{table}[t]
\caption{Top 8 topics (measured by size) of the ACL-IMDb dataset using either the censored or uncensored representation. The words in bold are sentiment related.}
\label{imdb_topics}
\vskip 0.15in
\begin{center}
\begin{small}
\addtolength{\tabcolsep}{-4pt}
\begin{tabular}{rll}
\toprule
  &                  Uncensored &                       Censored \\
\midrule
    1 &         acting, \textbf{bad}, script & acting, cinematography, actors \\
    2 &  role, actress, performance &            movie, watch, \textbf{waste} \\
    3 &           film, images, art &       actor, cast, performance \\
    4 &           \textbf{worst}, \textbf{waste}, \textbf{bad} &      saw, watched, documentary \\
    5 &            book, \textbf{felt}, read &            horror, scary, gore \\
    6 & \textbf{great}, \textbf{recommend}, \textbf{excellent} &          action, fight, movies \\
    7 &        \textbf{funny}, comedy, \textbf{laugh} &           \textbf{funny}, \textbf{laugh}, comedy \\
    8 &           police, gang, cop &      musical, dancing, actress \\
\bottomrule
\end{tabular}
\end{small}
\end{center}
\vskip -0.1in
\end{table}
The words in bold are related to sentiment. We can see that in the uncensored representations, we have several topics which are centered around sentiment (``worst, waste, bad'' and ``great, recommend, excellent ''), while our representations are fairly objective apart from a few sentiment-related words. Thus, if we want to find topics using a prelearned representation which are not biased by sentiment, it is important to censor the representations first.

\section{Results}
\label{results}
\begin{figure}[ht]
\vskip 0.2in
\begin{center}
\centerline{\includegraphics[width=\columnwidth]{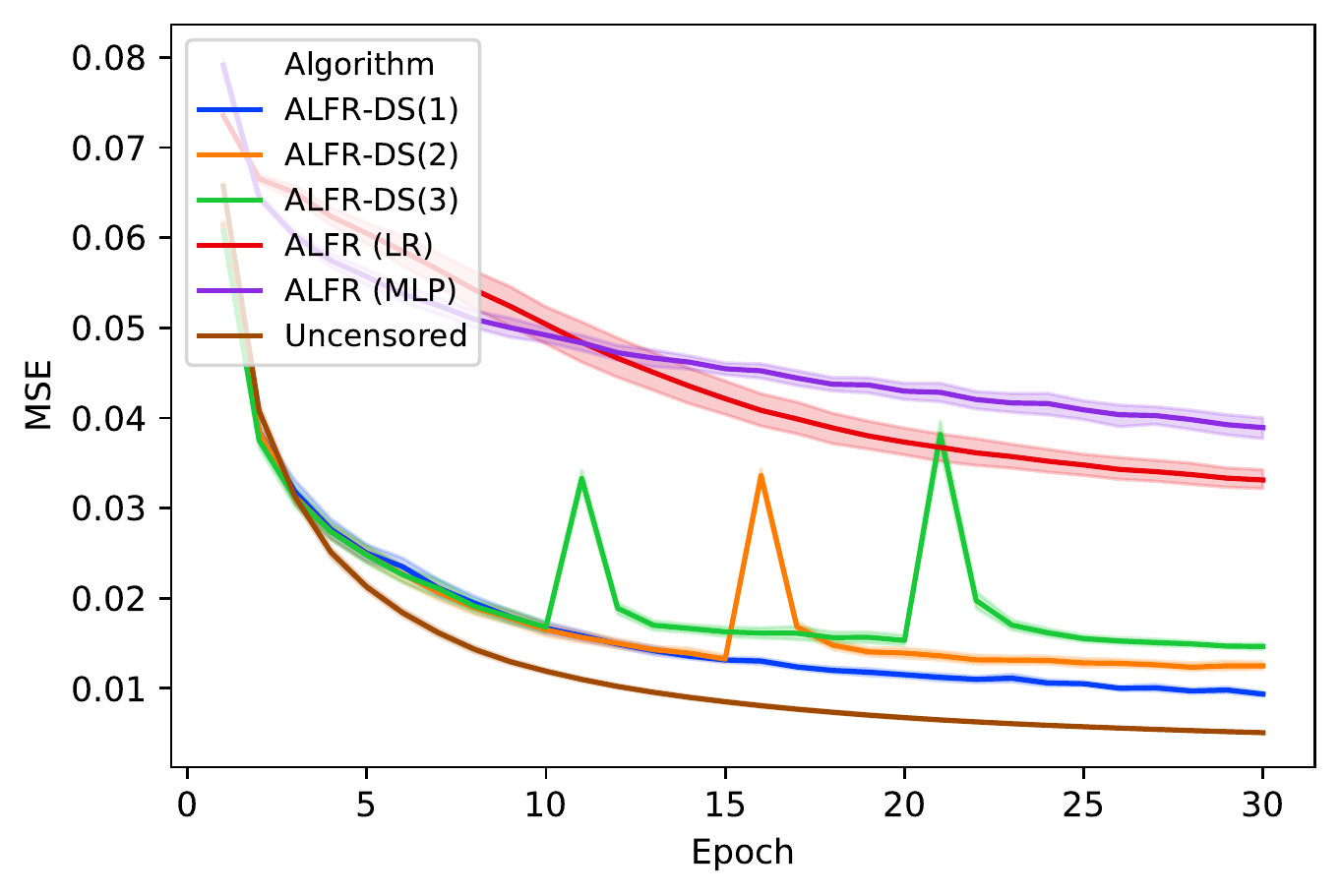}}
\caption{Reconstruction error (MSE) over 30 epochs for the MNIST task.}
\label{loss_mnist}
\end{center}
\vskip -0.2in
\end{figure}

\begin{figure}[ht]
\vskip 0.2in
\begin{center}
\centerline{\includegraphics[width=\columnwidth]{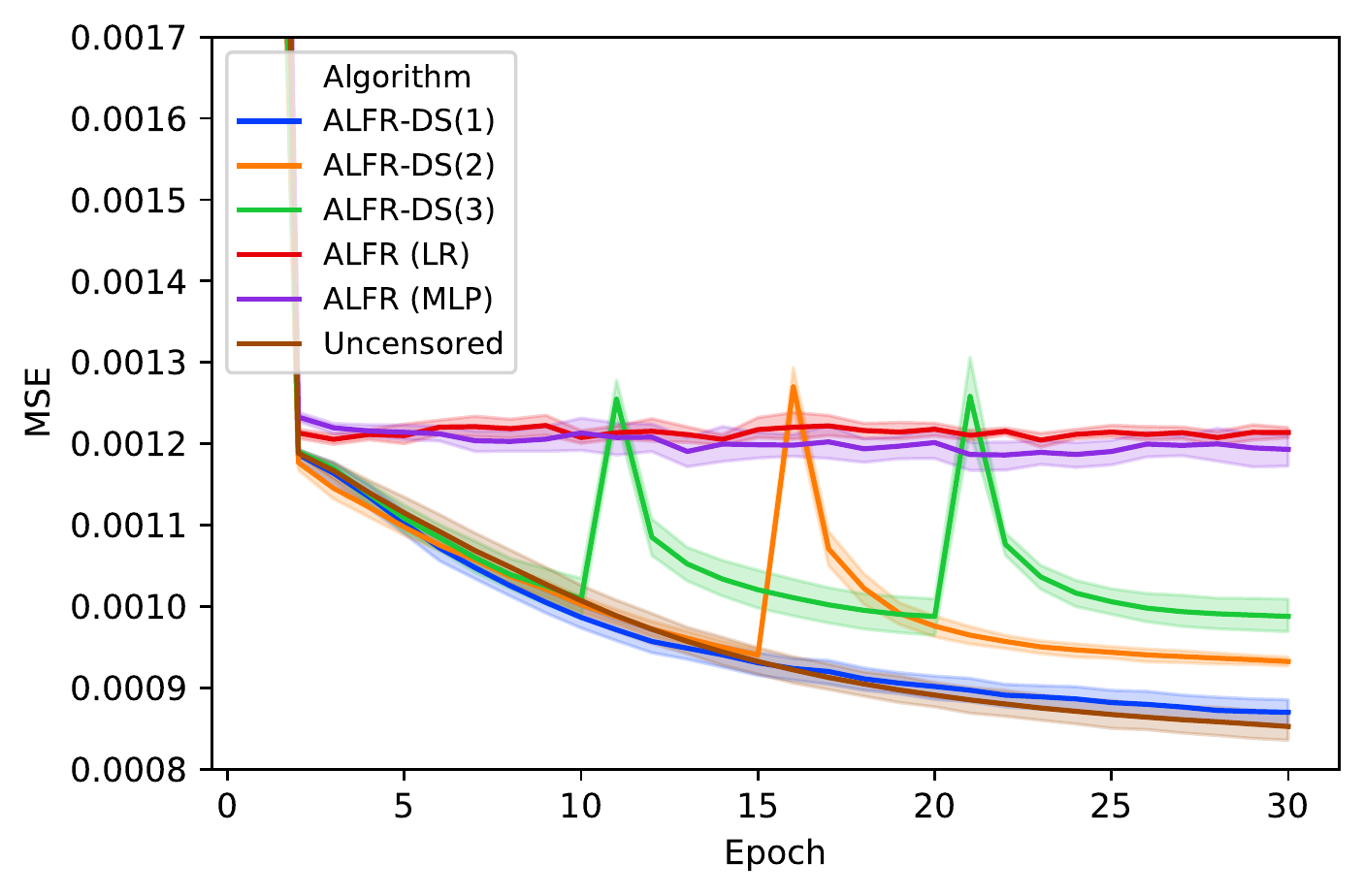}}
\caption{Reconstruction error (MSE) over 30 epochs for the ACL-IMDb task.}
\label{loss_imdb}
\end{center}
\vskip -0.2in
\end{figure}

As previously mentioned, the aim is to find censored representations with low reconstruction error and for which adversaries receive low accuracy scores for predicting $S$. In Figure~\ref{loss_mnist} and Figure~\ref{loss_imdb} we can see the reconstruction error for ALFR-DS, ALFR and Uncensored for the MNIST and ACL-IMDb task respectively during training time. Final reconstruction errors on a separate test set are not reported since they are equivalent of the final reconstruction error on the train set.

It is clear that ALFR-DS is much better at preserving the information of the original representations, even often staying very close to the uncensored representation. The ``spikes'' in the graph occur at the times in which a new encoder is added to the stack; in this case the loss briefly rises since each new encoder is initiated randomly. What is also interesting to note is that for the first 3 to 5 epochs, the graphs of ALFR-DS and Uncensored are almost identical. This is due to the fact that especially in the early phases dampening will be relatively small: the learned representations are still evolving a lot, meaning that the adversary has no time to become competent at predicting $S$. Only when the representations start to stabilize after approximately 5 epochs, we start to see a difference in the graphs.

The results for the final accuracy scores once a censored representation is learned is reported in Table~\ref{accuracy_scores}.
\begin{table}[t]
\label{accuracy_scores}
\caption{All the accuracy scores of different adversaries with increasing strength trained on the final representation as learned by ALFR-DS and ALFR. Lower is better, and best is highlighted in bold.}
\label{sample-table}
\vskip 0.15in
\begin{center}
\begin{small}
\addtolength{\tabcolsep}{-2pt}
\begin{tabular}{l|ll|ll}
\toprule
{} & \multicolumn{2}{c|}{\textbf{MNIST}} & \multicolumn{2}{c}{\textbf{ACL-IMDb}} \\
\midrule
{} & \multicolumn{1}{c}{\textbf{LR}} & \multicolumn{1}{c|}{\textbf{MLP}} & \multicolumn{1}{c}{\textbf{LR}} & \multicolumn{1}{c}{\textbf{MLP}} \\
\midrule
\tiny{ALFR-DS(1)} &  0.91 ± 0.01 &  0.98 ± 0.01 &  0.58 ± 0.03 &  0.61 ± 0.04 \\
\tiny{ALFR-DS(2)} &    \textbf{0.9 ± 0.0} &  0.95 ± 0.01 &  \textbf{0.53 ± 0.02} &  0.54 ± 0.02 \\
\tiny{ALFR-DS(3)} &   0.91 ± 0.0 &  \textbf{0.93 ± 0.02} &  0.53 ± 0.03 &  \textbf{0.53 ± 0.03} \\
\tiny{ALFR (LR)}  &  0.92 ± 0.03 &  0.95 ± 0.01 &  0.62 ± 0.14 &  0.65 ± 0.13 \\
\tiny{ALFR (MLP)} &  0.93 ± 0.02 &  0.95 ± 0.01 &  0.68 ± 0.12 &  0.69 ± 0.12 \\
\tiny{Uncensored} &   0.96 ± 0.0 &   0.99 ± 0.0 &  0.79 ± 0.01 &   0.81 ± 0.0 \\
\bottomrule
\end{tabular}
\end{small}
\end{center}
\vskip -0.1in
\end{table}
The lowest theoretical accuracy that can be achieved due to the distribution of $S$ is $0.9$ for the MNIST task, and $0.5$ for the ACL-IMDb task. The highest possible accuracies are reported for Uncensored. Again ALFR-DS achieves superior results in both censoring and reconstructing over ALFR. Specifically with respect to the ACL-IMDb task, we see that ALFR has much more variation in the reported accuracies in comparison to ALFR-DS, which is due to the unstable nature of ALFR. It is clear from the results that the number of stacked encoders in ALFR-DS does come at the cost of reconstruction. Particularly, ALFR-DS(1) in both tasks is greatly outperformed by ALFR-DS(3) in terms of censoring, however ALFR-DS(1) leaves much more of the data intact. It is thus clear that a balance has to be struck between censoring and reconstruction, which is normally handled by the algorithm with the use of the $\mathop{score}$ function with accompanying threshold $T$.
In both tasks when it comes to censoring, ALFR-DS(2) and ALFR-DS(3) perform similarly except when a strong nonlinear MLP adversary was tasked to predict $S$, in which case it becomes clear that ALFR-DS(3) has greater censoring capabilities. It is noteworthy to observe that ALFR(LR) outperforms ALFR(MLP) for censoring even though a weaker adversary was used during training. The reason is that LR converges much faster, and thus gives a more informative loss signal to the actor after each turn. Due to dampening, ALFR-DS does not have this problem and can reliably be used against strong adversaries.

As already mentioned, we have only reported results for ALFR when $\alpha=1$. We have found that ALFR-DS outperforms ALFR in both censoring and reconstruction for almost all values of $\alpha$, except for the cases when ALFR learns the trivial functions, i.e. an uncensored encoder or a constant function.
\section{Conclusions}
\label{conclusions}
In this paper, we have given a novel algorithm that uses dampening to stabilize the interaction between actor and adversary, and uses stacking to learn strong censored representations within a restricted hypothesis space. This algorithm outperforms the current approach in both censoring and reconstruction, as shown in our empirical results.

It should be noted that in this adversarial approach, the censoring capabilities are highly dependent on the strength of the adversary we train against. If we censor against a weak opponent, a strong opponent might still be able to uncover some of the unwanted bias. We believe it would be interesting to consider a setting where we train against multiple adversaries in increasing complexity and strength in order to decide what the optimal trade-off between reconstruction and censoring is.

Another line for future research is to address the increasing complexity of the encoder during training. It is often unwanted for reasons of efficiency to use large models. Moreover, such a model may contain a lot of redundancy, since the completion of the inner loop of our algorithm does not guarantee that the encoder is always optimally converged. Perhaps recent advances in Neural Network Compression can assist with this.

\bibliography{paper}
\bibliographystyle{icml2022}
\end{document}